\documentclass[conference]{IEEEtran}
\IEEEoverridecommandlockouts
\usepackage{cite}
\usepackage{amsmath,amssymb,amsfonts}
\usepackage{graphicx}
\usepackage{textcomp}
\usepackage{xcolor}
\usepackage{relsize}
\usepackage{xspace}
\usepackage{algorithm,algpseudocode}
\usepackage{comment}
\usepackage{multirow}
\newcommand{\argmin}[1]{\underset{#1}{\operatorname{arg\;min}}\;}
\usepackage{tikz}
\def\checkmark{\tikz\fill[scale=0.4](0,.35) -- (.25,0) -- (1,.7) -- (.25,.15) -- cycle;} 

\usepackage{subcaption}
\newcommand{\algo}{DisCERN\xspace}
\newcommand{\Break}{\textbf{Break} }
\begin{document}

\title{DisCERN:Discovering Counterfactual Explanations using Relevance Features from Neighbourhoods\\
\thanks{This research is funded by the iSee project~(https://isee4xai.com) which received funding from EPSRC under the grant number EP/V061755/1. iSee is part of the CHIST-ERA pathfinder programme for European coordinated research on future and emerging information and communication technologies.
}
}

\author{\IEEEauthorblockN{Nirmalie Wiratunga, Anjana Wijekoon, Ikechukwu Nkisi-Orji, Kyle Martin, Chamath Palihawadana, David Corsar}
\IEEEauthorblockA{\textit{School of Computing} \\
\textit{Robert Gordon University}\\
Aberdeen, Scotland \\
\{n.wiratunga, a.wijekoon1, i.nkisi-orji, k.martin3, c.palihawadana, d.corsar1\}@rgu.ac.uk}
}

\maketitle

\begin{abstract}
Counterfactual explanations focus on ``actionable knowledge'' to help end-users understand how a machine learning outcome could be changed to a more desirable outcome. 
For this purpose a counterfactual explainer needs to discover input dependencies that relate to outcome changes. 
Identifying the minimum subset of feature changes needed to action an output change in the decision is an interesting challenge for counterfactual explainers. 
The DisCERN algorithm introduced in this paper is a case-based counter-factual explainer. 
Here counterfactuals are formed by replacing feature values from a nearest unlike neighbour (NUN) until an actionable change is observed. 
We show how widely adopted feature relevance-based explainers (i.e.  LIME, SHAP), can inform DisCERN to identify the minimum subset of ``actionable features''.
We demonstrate our DisCERN algorithm on five datasets in a comparative study with the widely used optimisation-based counterfactual approach DiCE. 
Our results demonstrate that DisCERN is an effective strategy to minimise actionable changes necessary to create good counterfactual explanations. 
\end{abstract}


\section{Introduction} \label{sec:intro}
Understanding a user's explanation need is central to a system's capability of provisioning an explanation which satisfies that need~\cite{Mohseni}. 
Typically an explanation generated by an AI system is considered to convey the internal state or workings of an algorithm that resulted in the system's decision~\cite{counterfactual}. 
In machine learning (ML) the decision tends to be a discrete label or class (or in the case of regression tasks a numeric value). 
Although explanations focused on the internal state or logic of the algorithm is helpful to ML researchers it is arguably less useful to an end-user who may be more interested in how their current circumstances could be changed to receive a desired (better) outcome in the future.  
This calls for explainable AI (XAI) methods that focus on discovering relationships between the input dependencies that led to the system's decision. 

Case-based reasoning (CBR) is a widely accepted methodology for problem-solving, where the use of ``similar problems to solve similar solutions'' promotes an inherently interpretable reasoning strategy~\cite{cbrproto}.
The neighbourhood of similar problems is commonly harnessed in explainable AI~\cite{taxonomy}. 
A Nearest-Like Neighbours (NLNs) based explainer, focuses on input dependencies by identifying similarity relationships between the current problem and the retrieved nearest neighbour~\cite{eoin}. 
Research has shown that when similarity computation is focused on feature selection~\cite{wiratunga} and weighting~\cite{wettschereck}, it can significantly improve retrieval of NLNs. 
The prevailing CBR approach to counterfactual generation harnesses similarity in neighbourhoods to identify similar cases with different class labels to the NLNs. 
Such neighbours, referred to as nearest unlike neighbours (NUNs), forms the basis for generating counterfactuals from neighbourhoods. 
Accordingly it would be reasonable to expect that NUN-based explanation generation would also benefit from the knowledge of feature importance. 
Certainly having to focus on a few key ``actionable'' features in domains with large numbers of features will be more desirable from a practical standpoint, as well as reducing the cognitive burden of understanding neighbourhood-based explanations.

Unlike NLN based explanations, a counterfactual explanation focuses on identifying ``actionable knowledge''; which is knowledge about important causal dependencies between the input and the ML decision. Such knowledge helps to understand what could be changed in the input to achieve a preferred (desired) decision outcome. 
Typically a NUN is used to identify the number of differences between the input and its neighbour, that when changed can lead to a change in the system's decision~\cite{goodcounterfactual}. 
A key challenge that we address in this paper is to identify the minimum subset of feature value changes to achieve that needed decision flip - the ``actionable features''. 
Accordingly the paper has the following contributions: 
\begin{itemize}
\item Discover the minimum actionable feature changes using feature relevance-based explainer methods like LIME~\cite{lime} or SHAP~\cite{shap}; 
\item Advances the case-based counterfactual generation research by introducing our \algo algorithm; and
\item Evidences the utility of \algo with results from a comparative study with multiple datasets and the optimisation-based DiCE~\cite{dice} counterfactual approach. 
\end{itemize}
The rest of the paper is organised as follows. Section~\ref{sec:related} investigates the importance of counterfactual XAI and discuss two key counterfactual methods and their evaluation methodologies. Feature Relevance Explainers are discussed and compared in Section~\ref{sec:fre} and  Section~\ref{sec:method} presents our \algo algorithm to improve the discovery of actionable features in counterfactuals. 
Section~\ref{sec:eval} presents the evaluation methodologies, datasets and performance metrics followed by results in Section~\ref{sec:results}. 
Finally we draw conclusions and discuss future work in Section~\ref{sec:conc}. 

\section{Related Work}
\label{sec:related}

Like many explanation methods, counterfactual explanations are rooted within the study of human psychology. Counterfactual thinking is a mental exercise where we attempt to isolate the specific actions which contributed to a (usually undesirable) outcome, with the goal of identifying how we could alter these facts to reach an alternative (and often more desirable) outcome~\cite{Roese}. In this manner we derive a form of causal explanation for the outcome that was actually achieved, allowing us to reason about how a different outcome could be achieved in future~\cite{Harris}. 
To clarify, consider the following fictitious example of a runner who was placed third in a race: ``I won the bronze medal for my race (actual outcome), but I would have won the gold medal (desired outcome) if I hadn't tripped (causal action which changed outcome)''. Through this thought process, the runner has derived what they believe to be a causal action that led to receiving the bronze medal. 
With that knowledge inferred, the runner can then reason that in order to achieve a better outcome, they should run a similar race again, but avoid tripping. 
Likewise with a counterfactual explanation we aim to explain why the system generated a specific outcome instead of another, by inferring important relationships (causal or otherwise) between input features~\cite{counterfactual}. 

Case-based Reasoning (CBR)~\cite{goodcounterfactual} and optimisation techniques~\cite{dice,yajioriginal} have been the pillars of discovering counterfactuals from data.  
Recent work in CBR has shown how counterfactual case generation can be conveniently supported through the case adaptation stage, where query-retrieval pairs of successful counterfactual explanation experiences are used to create an explanation case-base~\cite{goodcounterfactual}.  
Unlike the CBR approach to counterfactual generation, DiCE~\cite{dice}  trains a generative model using gradient descent optimisation to output multiple perturbed diverse counterfactuals.   
This optimisation upholds two key requirements of a good counterfactual which are: maximising the probability of obtaining the desired class~(i.e. discovered counterfactual class is different from query class); and minimising the distance to the query~(similar to discovering NUN).
Additionally the DiCE optimisation also maximises the distance between multiple counterfactuals to ensure that they are diverse. 
With the CBR approach additional counterfactuals can be identified by increasing the neighbourhood.
This ability to provide multiple counterfactuals to end-users has been found to improve end-user's mental model.
In our work we also adopt CBR's NUN method to find counterfactuals but instead of the adaptation CBR step or the DiCE optimisation, 
we opt for feature relevance explainers to inform us about actionable features. 
In doing so we avoid the need to create similarity-based explanation case-bases, yet maintain the advantage of locality-based explanations which ensure valid counterfactuals that are often harder to guarantee with optimisation methods.  

Discussions regarding user acceptability and satisfiability of explanations has dominated XAI research in recent years~\cite{Hoffman2018MetricsFE}.
Quantitative evaluation of counterfactual explanations focus on measures that can ascertain properties of good counterfactuals. 
In case-based counterfactual research, an explainer CBR system's ability to solve future ``explanation queries'' is measured by explanatory competency~\cite{goodcounterfactual}.
This measures the fraction of queries that are currently explained by the explainer CBR system - coverage of the counterfactual cases.
Authors claim that a good counterfactual will have at most two feature-differences (although this is not guaranteed by the explainer CBR system) and thereby maintain minimum plausible changes by operating within a local neighbourhood.
Other related work also confirm the importance of measuring nearness and minimal changes to evaluate counterfactual explanations through proximity and sparsity measures~\cite{dice}. 
Measures such as validity and diversity are also used with generative counterfactual XAI methods. 
However these are not applicable to our work, as case-based counterfactuals have the advantage of formulating plausible feature changes using valid cases in the case-base. 
Accordingly in our evaluation strategy, we use proximity and sparsity to compare counterfactual methods and also measure efficiency of actionable feature discovery. 

\section{Feature Relevance Explainers}\label{sec:fre}
Feature relevance weights conveys the extent to which a feature contributes to a model's output - i.e. greater weight indicates greater relevance of that feature to model decision-making. 
Features with the largest weights can therefore be used to explain the contributory input values that resulted in the output decision.
In \algo, relevance weights are used to partially order features for actionable feature discovery. 
In the rest of this section we discuss and compare two widely used feature relevance explainers; LIME and SHAP.

\subsection{Local Interpretable Model-agnostic Explanations (LIME)}\label{lime}
LIME~\cite{lime} is a model-agnostic feature relevance explainer which creates an interpretable model around a data instance to estimate how each feature contributed to the black-box model outcome.
LIME creates a set of perturbations within the instance's neighbourhood and labels them using the black-box model. This newly labelled dataset is used to create a linear interpretable model (e.g. a weighted linear regression model). 
The resulting surrogate model is interpretable and only locally faithful to the black-box model~(i.e. correctly classifies the input instance, but not all data instances outside its immediate neighbourhood). The new interpretable model is used to predict the classification outcome of for the data instance.
Thereafter an explanation of the predicted class is formed by obtaining the weights that indicate how each feature contributed to the outcome. 

\subsection{SHapley Additive exPlanation (SHAP)}\label{shap}
SHAP~\cite{shap} is a model-agnostic feature relevance explainer with theoretical guarantees about consistency and local accuracy from game theory and based on the shapley regression values~\cite{shapley}.
Shapley values are calculated by creating linear models using subsets of features present in a case-base, $X$ (i.e. a set of data instances). 
More specifically, a model is trained with a subset of features of size, $m'$, and another model is trained with a subset of features of size, $m'+\hat{m}$. 
Here, $m'+\hat{m} <= m$, and the second model additionally includes a set of features, $\hat{m}$, selected from the set of features that were left out in the first model. 
A set of such model pairs are created for all possible feature combinations. 
For a given data instance that needs to be explained, the prediction differences of these model pairs are averaged to find the explainable feature relevance weights.

\subsection{Feature Relevance Explainer Properties}

LIME and SHAP are Post-hoc, model-agnostic feature relevance explainers (see Table~\ref{tbl:fre}). 
Shapley values are considered to be \emph{consistent} (i.e the same query results in the same relevance explanation), 
unlike LIME which discovers relevance weights using perturbed data. 
Both SHAP and Lime guarantees local \emph{accuracy} (i.e. the surrogate model and black-box model predicts the same outcome for a data instance); and \emph{missingness} (i.e. ensures that there there is no weight contribution from a missing feature). 
Both provide a feature relevance weights vector, for any given query. 
In our counterfactual work the magnitude of the relevance weights, focus the search for the minimal subset of feature value changes that are likely to bring about a class change for a given query. 

\begin{table}[t]
\centering
\caption{Comparison of Feature Relevance Explainers}
\renewcommand{\arraystretch}{1.2}
\begin{tabular} {l@{\hspace{10pt}}r@{\hspace{5pt}}r}
\hline
Property&LIME&SHAP\\
\hline
Explainer Type&Post-hoc&Post-hoc\\
Model dependency&Model-agnostic&Model-agnostic\\
Explainability&Linear &Game Theory\\
Principal&Approximation&Inspired\\
Local Accuracy&\checkmark&\checkmark\\
Missingness&\checkmark&\checkmark\\
Consistency& - &\checkmark\\
\hline
\end{tabular}
\label{tbl:fre}
\end{table}

\section{Methods}
\label{sec:method}
\algo algorithm uses feature relevance to identify the minimum subset of changes needed to form a counterfactual explanation from a retrieved NUN.
Here we formalise the NUN counterfactual approach and thereafter discuss how weights from, feature relevance explainers, can be used in \algo to discover minimal number of actionable features. 

\subsection{Nearest-Unlike-Neighbour Counterfactual}\label{sec:nun}
\begin{figure}[t]
\centering
\includegraphics[width=0.3\textwidth]{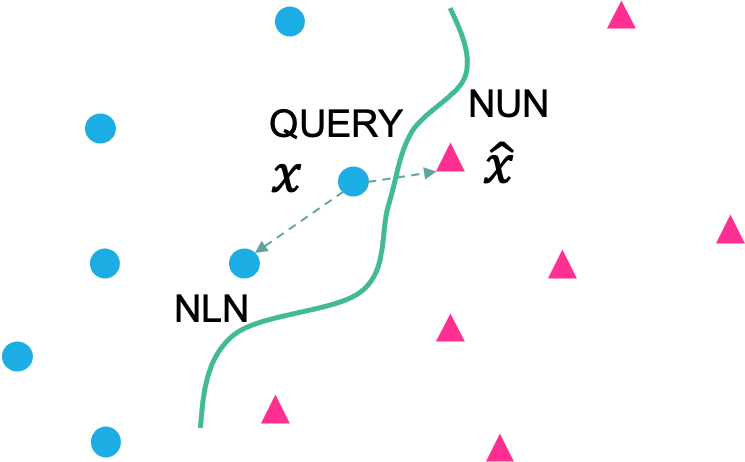}
\caption{Nearest Like and Unlike Neighbours in 2D space} 
\label{fig:nun}
\end{figure}

The goal of a counterfactual explanation is to guide the end-user to achieve class change~(i.e. actionable), with a focus on minimising the number of changes needed to flip the decision to a more desirable outcome.
Given a query instance, $x = \{x_1, x_2, ... , x_m\}$, with $m$ features, 
its counterfactual, $\hat{x} = \{\hat{x}_1, \hat{x}_2, ..., \hat{x}_m\}$, is identified as the NUN in the case-base, $X$~\cite{counterfactual}, as follows: 
\begin{equation}
\begin{gathered}
    \hat{x} \leftarrow \argmin{x' \in X} d(x, x');\;x, x', \hat{x} \in X;\; y \neq y'
\end{gathered}
\label{eq:nun}
\end{equation}
Figure~\ref{fig:nun} illustrates a neighbourhood for a binary classed problem in two dimensional feature space~($m=2$).
It shows for a given query, how a NUN appears close to its decision boundary.
In theory the closer it is to the boundary the fewer actionable changes are likely to be needed to flip the decision. 
However in practice certain types of feature changes (even if small) may be harder, or in some cases, even impossible to action (e.g. features related to demographics). 
Such discoveries may still be useful to unearth, because they can point to unethical or unfair practices. 

Paired distances between the query and candidate NUNs (i.e. other data instances with a different class to that of the query) are computed 
using Euclidean distance~(Equation~\ref{eq:ecd}). 
\begin{equation}
d(x, \hat{x}) = \sqrt{\mathlarger{\sum_{i=1}^m} (x_i - \hat{x}_i)^2}
\label{eq:ecd}
\end{equation}
The optimal NUN will have the smallest distance to the query.
Both the query's and NUN's feature values are used by \algo to formulate the counterfactual explanation. 
Here we identify two challenges: discovering the minimal number of actionable features (from a maximum of $m$ potential feature changes) and minimising the amount of change required for each feature. We address these challenges using relevance weights.

\subsection{Feature weights from a Feature Relevance Explainer}

\begin{figure*}[t]
\centering
\begin{subfigure}{0.32\textwidth}
\centering
\includegraphics[width=1\textwidth]{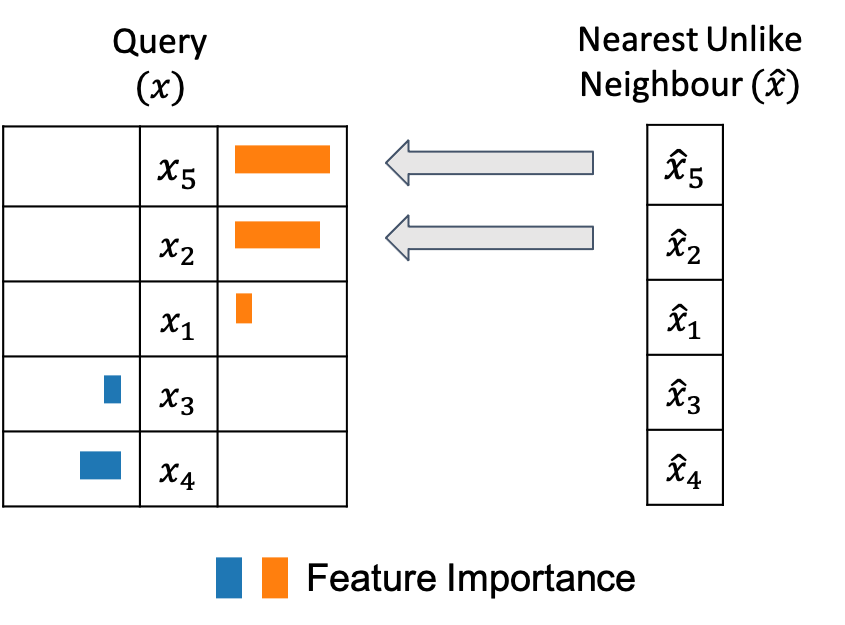}
\caption{Query Relevance (QRel)}
\label{fig:qi}
\end{subfigure}
\begin{subfigure}{0.32\textwidth}
\centering
\includegraphics[width=1\textwidth]{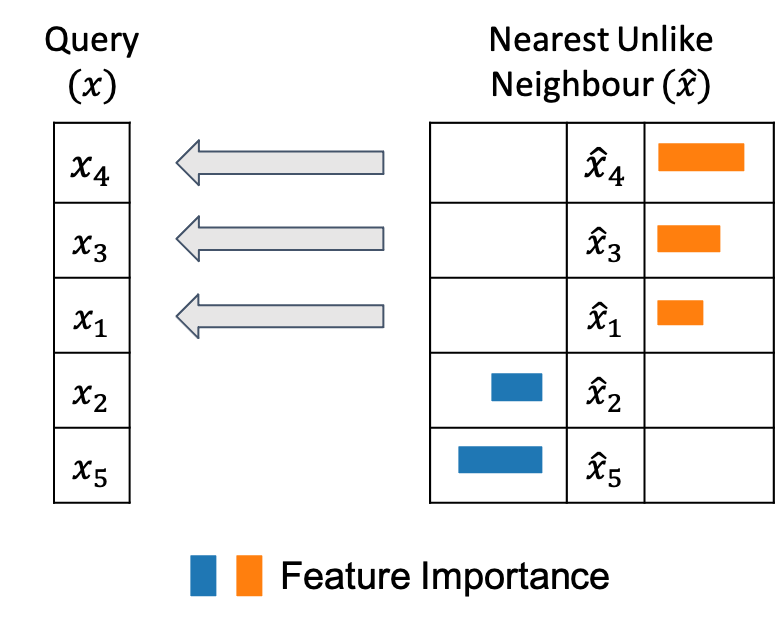}
\caption{NUN Relevance (NRel)}
\label{fig:ni}
\end{subfigure}
\caption{Actionable Feature Discovery with QRel and NRel}
\label{fig:methods}
\end{figure*}
For an arbitrary data instance, $p$, the output of a feature relevance explainer, $Rel$, is a list of weights, $\mathbf{w}=(w_1, w_2,\ldots, w_m)$, where $w_i$ is a real-valued weight assigned to feature, $p_i$:
\begin{equation}
    Rel(p) \rightarrow \{\mathbf{w}\;|\;\mathbf{w}\in \mathbb{R}^m \}
\label{eq:fw}
\end{equation}
A positive weight~($w_i >= 0$) indicates that the corresponding feature contributes positively and a negative weight~($w_i < 0$) contributes negatively towards the predicted outcome. 
These relevance weights are used to define an ordering constraint $\mathcal{R}$ on features. Given a weights vector, $\mathbf{w}$, the overall value, $w(.)$, is the weight lookup of a feature included in $p$. This is used to define the ordering $\preceq$ on features:
\begin{equation}
     p_i \preceq_\mathcal{R} p_j \iff \mathcal{R}:: w(p_i) \geq w(p_j)
\label{eq:fwx}
\end{equation}

\subsection{Actionable Feature Discovery with Feature Relevance Explainers}
\label{sec:afd}
The number of feature changes, $n$, required to achieve a class change, can range from 1 to $m$ ($1\leq$$n$$\leq$$m$). 
We propose two methods to discover actionable features, with the goal of minimising the number of feature changes~($n$) needed to form a counterfactual explanation.
The first method replaces values of the most important features in the query with the corresponding feature values from its NUN; or a second alternative is to identify the most relevant features of the NUN and copy those feature values to modify the query instance.

Figure~\ref{fig:methods} shows a query with five features being adapted to form a counterfactual using two alternative value replacement methods: QRel in Figure~\ref{fig:qi}; and NRel in Figure~\ref{fig:ni}.
With both methods, values are copied from the NUN, 
but the exact features selected for replacement change based on the instance parameter used in $Rel(.)$.
In QRel, the query features are ordered by their feature weights and the most important features are replaced by the respective NUN feature values. 
That is, QRel, uses $Rel(p=x)$ to obtain feature weights. 
With NRel, the NUN features are ordered by their feature weights returned by $Rel(p=\hat{x})$, 
and the most important features are reused by the query. 
Since each method imposes a different feature ordering, the number of changes needed to form the final counterfactual is likely to be different. 
In our example QRel and NRel achieve class change with 2 and 3 feature replacements respectively in Figure~\ref{fig:methods}. 

\subsection{\algo Algorithm}\label{discern}

Algorithm~\ref{algo:afd} brings together methods from Sections~\ref{sec:nun} through~\ref{sec:afd} to form NUN counterfactuals.
Here, $y=\mathcal{F}(x)$, is the (black-box) classifier prediction for the query and, $\hat{y}=\mathcal{F}(\hat{x})$, is the class prediction for the NUN.
Relevance weights from Section~\ref{sec:afd} are identified in reference to either the query or the NUN, which we have denoted as $p$; and the $Order$ method (from Equation~\ref{eq:fwx}) provides a list of feature indices ranked by the relevance weights.
Feature values are iteratively replaced until the actionable change condition is met (i.e., $\mathcal{F}(x')=y' \wedge  y \neq y'$) or the query is completely changed to the NUN (which is guaranteed to result in class change). 
Here the fewer the iterations needed the better the actionable features discovered. 
Once class change is achieved 
DisCERN returns $x'$ as the counterfactual.

\begin{algorithm}[t]
\caption{\algo[$Rel$,\;$p$]}
\label{algo:afd}
\begin{algorithmic}[1]
\Require $(x,y)$: query and label pair
\Require $(\hat{x}, \hat{y})$: NUN and label pair
\Require $\mathcal{F}$: classification model
\Require $p$: either $x$ or $\hat{x}$ \Comment{as described in Section~\ref{sec:afd}}
\Require $Rel$: Feature Relevance Explainer
\State $\mathbf{w} = Rel(p)$\Comment{see Equation~\ref{eq:fw}} 
\State $\mathbf{\hat{w}} = Order(\mathbf{w})$\Comment{see partial order Equation~\ref{eq:fwx}}
\State Initialise $y' = y$; $x' = x$\Comment{init counterfactual as query} 
\For {$w_i \in \mathbf{\hat{w}}$} 
    \If {$x'_i \neq \hat{x}_i$} \Comment{$i$ is the index of $w_i$}
        \State $x'_i \leftarrow \hat{x}_i$ \Comment{copy NUN feature value}
        \State $y' = \mathcal{F}(x')$ \Comment{class prediction for counterfactual}
        \If  {$y' \neq y$}
            \State \Break
        \EndIf
    \EndIf
\EndFor
\State
\Return $x'$
\end{algorithmic}
\end{algorithm}

\section{Evaluation}
\label{sec:eval}

The goal of this evaluation is twofold. 
Firstly to investigate different relevance feature explainers discussed in Section~\ref{sec:fre} with a view to finding which is best for feature weighting with \algo.
Secondly to conduct a comparative study to determine the effectiveness of the \algo counterfactual explanation method (in Section~\ref{discern}) 
with the widely-used DiCE~\cite{dice} and the baseline random feature ordered counterfactual creation methods.

Since \algo is a combination of a feature relevance explainer, $Rel$, and an actionable feature discovery method based on $Rel(p=x)$ or $Rel(p=\hat{x})$; 
we use the abbreviated notation \algo[$Rel$,\;$p$] to refer to these configuration combinations,
where
$Rel\leftarrow$\text{(RND$\vert$LIME$\vert$SHAP$\vert$LIME}$_C\vert$\text{SHAP}$_C$\text{)}; and 
$p\leftarrow$\text{(QRel$\vert$NRel)}.
For example \algo[LIME,\;QRel] denotes the combination of LIME weights with the QRel actionable feature discovery method for counterfactual creation. 
Note that RND refers to random selection of actionable features and therefore 
does not depend on either the query or NUN for parameter $p$, and is denoted as \algo[RND, Null].


\subsection{Datasets}\label{sec:data}
Datasets used in this paper are summarised in Table~\ref{tbl:data},
and used to carry out experiments as follows: 
\begin{itemize}
\item comparative study of relevance feature explainers for feature weighting uses the Moodle dataset with results analysed in Section~\ref{sec:res-fos}; 
and 
\item the counterfactual creation experiment is conducted using the Moodle dataset and four further datasets: Loan-2015, Alcohol, Income and Credit datasets. These results appear in Section~\ref{sec:res-afd}. 
\end{itemize}
Accuracy for each dataset is reported using a RandomForest classifier of 500 trees, which we found had better performance over several other black box models (including neural nets) in an initial set of fine-tuning experiments (with stratified 3 fold validation)
For convenience with explanation experiments, we assume that the black box model has correctly predicted the outcome, and hence it made sense to work with the model with highest dataset accuracy.
In this paper all features are considered candidates for actionable features\footnote{ 
In practice this is unlikely to be always possible. For instance features such as income, home ownership and length of employment in the Loan-2015 dataset are naturally non-actionable; and so are demographic features.}. 
Details of how each dataset is preprocessed and used in a counterfactual explanation scenario are discussed next. 

\begin{table}[t]
\centering
\caption{Dataset Properties}
\renewcommand{\arraystretch}{1.2}
\begin{tabular} {l@{\hspace{10pt}}r@{\hspace{5pt}}r@{\hspace{5pt}}r@{\hspace{5pt}}r@{\hspace{5pt}}r@{\hspace{5pt}}r}
\hline
Datasets&Moodle&Loan-2015&Alcohol&Income&Credit\\
\hline
Features&95&77&5&15&15\\
Continuous Features&95&68&3&7&6\\
Categorical Features&0&9&2&8&9\\
Classes&2&2&2&2&2\\
Classification Accuracy&83\%&97\%&99\%&84\%&77\%\\
\hline
\end{tabular}
\label{tbl:data}
\end{table}
 
\subsubsection{Moodle Dataset} is constructed from records of student footprints on the Moodle Virtual Learning Environment (VLE) for a single class delivered within Robert Gordon University, UK (RGU). VLE interactions help to capture vital touchpoints that can be used as proxy measures of student engagement. 
The dataset consists of 74 students enrolled on a Computer Science class during Semester 1 of 2020/2021 at RGU. It contains 95 features, where each feature is a learning resource stored on the Moodle VLE and the feature value is the number of times it was accessed by a student. 

The ML task is to predict if a student gets a higher or a lower grade based on their Moodle footprint. This task is based on the assumption that there is some (causal or other) relationship between the Moodle access and the final grade of a student. 
We consider grades \textit{A} and \textit{B} as \textit{Higher} grades and \textit{C}, \textit{D}, \textit{E} and \textit{F} as \textit{Lower} grades. Grades were consolidated as \textit{Higher} and \textit{Lower} to mitigate the comparably lower number of data instances and class imbalance. 
This formed a dataset of 74 instances for a binary grade classification task, based on the Moodle footprint. 
The explanation intent relevant to this dataset is of the type \textit{Why} student A did \textit{not} receive a higher grade X? instead of \textit{Why} did student A receive grade Y? 
The latter can be explained using a feature relevance explainer presenting the contribution of the most important features for the predicted grade; and the former \textit{Why not} type questions are better explained through a counterfactual explanation with actionable features to guide the student to achieve a more desirable outcome in the future. 

\subsubsection{Loan-2015 dataset}
is the subset of 2015 records from the Lending Club loan dataset on kaggle\footnote{https://www.kaggle.com/wordsforthewise/lending-club}. 
We limit the dataset to records from 2015 to create the loan-2015 dataset of 421,095 data instances with 151 features. 
The ML task is to predict if a loan will be paid in full or not, and this outcome is used to accept or reject future loan requests. We apply data pre-processing steps recommended by the data providers to obtain a dataset with 342,865 instances and 115 features to perform binary classification. 
The desirable outcome for an end-user is that his/her loan request is accepted~(i.e. similar users successfully re-paid their loans). 
For example an explanation intent here can trigger a question such as \textit{Why} person A did \textit{not} receive the loan? with a counterfactual pointing to those features in need of adjustments, before a desirable outcome is possible in the future. 

\subsubsection{Alcohol Dataset} is the Blood Alcohol Concentration~(BAC) dataset which consists of 127,800 data instances with 7 features~\cite{BAC}. It includes features such as gender, if a meal was taken, the duration between the meal and BAC test. 
The ML task for this dataset is to predict if the BAC is over a regulatory limit. Accordingly, in a pre-processing step the dataset is converted in to a binary classification task by recognising the two classes with the BAC regulatory limit as the decision threshold. 
In common with the previous two datasets, one of the outcomes is more desirable than the other; 
which is of having the BAC below the acceptable threshold for driving. 
Accordingly, counterfactual explanations are sought by individuals who have a BAC above the threshold and are looking to understand how they might keep their BAC below the threshold by better managing one or more actionable features (e.g. such as periods between meal and alcohol consumption). 

\subsubsection{Income Dataset} contains 45222 data instances of personal data based on US 1994 Census database\footnote{https://archive.ics.uci.edu/ml/datasets/adult}. There are 15 features including demographic, educational, and other personal properties to predict their yearly income in US Dollars. 
The ML task for this data set is to predict, if the income of a person is above or below 50k per year. Accordingly, in a pre-processing step the dataset is converted in to a binary classification task by setting the two classes to be less than or equal to 50k, and above 50k; with the latter being the desirable class. 
A counterfactual explanation is sought out by an individual with a salary below 50k and seeking to change one or more of their circumstances (such as their educational or demographic attributes) to acheive the higher salary class. 

\subsubsection{Credit Dataset} is a credit card application approval dataset with 653 data instances\footnote{https://archive.ics.uci.edu/ml/datasets/Credit+Approval}. There are 15 anonymised features describing an applicant with the class indicating if the credit card application was approved or not. 
In this binary classification task the the model predicts if an applicant should be approved or not. 
For an applicant the desirable outcome is an ``approved'' credit card application. 
Accordingly, applicants who seek counterfactual explanations would have typically ``failed'' their credit card application and are looking to change this to an ``approved'' outcome by identifying necessary feature changes. 

\subsection{Performance Measures}

Four quantitative performance measures (validity, proximity, sparsity and diversity) for evaluating counterfactuals were introduced in~\cite{dice}. Proximity measures the mean feature-wise distance between a query and its counterfactual explanation. Sparsity refers to the number of features that are different between a query and its counterfactual explanation. Validity measures if the counterfactuals presented by the method actually belongs to the desirable class~(i.e. not the same class as the query). 
Diversity measures the heterogeneity between multiple counterfactuals. 
Both validity and diversity are not relevant for our work because: 
1) by selecting a NUN we ensure 100\% validity; and 
2) by selecting the most similar NUN counterfactual there is no requirement to measure diversity. 
In this paper we present two measures which correspond to sparsity and proximity respectively, but are not identical to the measures in~\cite{dice}. 

\subsubsection{Mean number of feature changes~($\#F$)} required to achieve class change is calculated as follows: 
\begin{equation}
    \#F = \dfrac{1}{N\times m}\sum_{j=1}^{N} \sum_{i=1}^{m} 1_{[\hat{x}_{i} \neq x_{i}]}
\end{equation}
Here the number of features with different values between the counterfactual~($\hat{x}$) and the query~($x$) are calculated and averaged; where $N$ refers to the number of query instances, and $m$ is the number of features. 

\subsubsection{Mean amount of feature changes~($\$F$)} required to achieve class change is calculated as follows:
\begin{equation}
    \$F = \dfrac{1}{N\times \#F}\sum_{j=1}^{N} \sum_{i=1}^{m} (|\hat{x}_{i} - x_{i}|)
\end{equation}
Here the sum of feature differences are average over $\#F$ and the number of query instances~($N$). All continuous features are min/max normalised and therefore, continuous feature differences are between 0 and 1 whereas categorical feature differences are always 1 (using the overlap distance). 
Accordingly, datasets with more categorical features will have higher $\$F$, which means that the $\$F$ measure is not comparable across datasets.
Note that smaller values of $\#F$ and $\$F$ are desirable. 

\section{Experimental Results}
\label{sec:results}
\subsection{Comparison of Weights from Relevance Feature Explainers}\label{sec:res-fos}
For the first study we compare the following relevance feature explainers for \algo:
\begin{enumerate}
\item LIME: feature weights from the local feature relevance explainer discussed in Section~\ref{lime}.
\item SHAP: feature weights provided by Shapley values discussed in Section~\ref{shap}.
\item Chi2: global feature weights from the Chi-Squared test for feature selection.
\item  LIME$_C$ and SHAP$_C$: these are two class level feature relevance explainer versions of LIME and SHAP weights respectively, where for each class, the aggregated feature relevance is the mean feature relevance weights over all train data instances for that class.
\end{enumerate}
\begin{table}[t]
\centering
\caption{Comparison of feature ordering strategies}
\renewcommand{\arraystretch}{1.2}
\begin{tabular} {l@{\hspace{6pt}}r@{\hspace{6pt}}r@{\hspace{16pt}}r@{\hspace{6pt}}r}
\hline
&\multicolumn{2}{c}{$\#F$}&\multicolumn{2}{c}{$\$F$}\\
\algo[Rel$\downarrow$,\; p$\rightarrow$]& QRel & NRel & QRel & NRel \\
\hline
LIME& 8.14& 8.61&0.2642&0.2726\\
LIME$_C$&11.41&10.38&0.2308&0.2524\\
SHAP&\textbf{7.69}&\textbf{8.32}&0.2660&0.2454\\
SHAP$_C$&10.28&10.18&\textbf{0.2085}&\textbf{0.2068}\\
Chi2&\multicolumn{2}{c}{12.27}&\multicolumn{2}{c}{0.2700}\\
\hline
\end{tabular}
\label{tbl:os}
\end{table}

\begin{table*}[t]
\centering
\caption{Comparison of counterfactual methods on $\#F$}
\renewcommand{\arraystretch}{1.2}
\begin{tabular} {l@{\hspace{16pt}}r@{\hspace{10pt}}r@{\hspace{10pt}}r@{\hspace{10pt}}r@{\hspace{10pt}}r@{\hspace{10pt}}r}
\hline
Datasets&Moodle&Loan-2015&Alcohol&Income&Credit\\
\hline
DiCE&10.21&\textbf{2.59}&2.53&2.95&2.81\\
\algo[RND,\;$Null$]&21.62&21.91&2.16&2.92&3.18\\
\algo[LIME,\;QRel]&8.14&6.93&\textbf{2.11}&\textbf{2.59}&\textbf{2.42}\\
\algo[LIME,\;NRel]&8.61&6.69&2.15&2.64&2.50\\
\algo[SHAP,\;QRel]&\textbf{7.69}&6.86&\textbf{2.11}&2.70&2.48\\
\algo[SHAP,\;NRel]&8.32&5.51&2.12&2.68&\textbf{2.42}\\
\hline
\end{tabular}
\label{tbl:sparsity}
\end{table*}

\begin{table*}[t]
\centering
\caption{Comparison of counterfactual methods on $\$F$}
\renewcommand{\arraystretch}{1.2}
\begin{tabular} {l@{\hspace{16pt}}r@{\hspace{10pt}}r@{\hspace{10pt}}r@{\hspace{10pt}}r@{\hspace{10pt}}r@{\hspace{10pt}}r}
\hline
Datasets&Moodle&Loan-2015&Alcohol&Income&Credit\\
\hline
DiCE&0.6344&0.7763&0.6707&0.8497&0.8179\\
\algo[RND,\;Null]&0.2924&\textbf{0.0569}&\textbf{0.0909}&0.3545&0.2573\\
\algo[LIME,\;QRel]&0.2642&0.0660&0.0927&0.3643&0.2365\\
\algo[LIME,\;NRel]&0.2726&0.0675&0.0910&0.3570&0.2662\\
\algo[SHAP,\;QRel]&0.2660&0.0760&0.0929&0.3604&0.2343\\
\algo[SHAP,\;NRel]&\textbf{0.2454}&0.0711&0.0925&\textbf{0.3258}&\textbf{0.2210}\\
\hline
\end{tabular}
\label{tbl:proximity}
\end{table*}
A comparison of \algo settings with 5 alternative options for $Rel$; and 2 alternatives for $p$ appear in Table~\ref{tbl:os} for the Moodle dataset.
Each alternative's performance is compared on $\#F$ and $\$F$ performance measures.
Note that there is no difference between QRel and NRel when using Chi2 because feature weightings are global and not determined by a local data instance (be that the query or the NUN). 
Results show that SHAP achieves the best performance over LIME and Chi2 with both QRel and NRel methods (see bold font). 
SHAP using the QRel feature ordering method has achieved lowest $\#F$, whilst SHAP$_C$ has the lowest $\$F$.
However, since SHAP$_C$ requires additional feature changes to achieve class change, we consider SHAP to be a preferable strategy. 
Moreover, we observe that LIME also achieves comparable performances for both $\#F$ and $\$F$. 
Notably, Chi2 failed to outperform both LIME and SHAP strategies in both minimising number of features and amount of change. Overall, these results emphasise the importance of feature relevance explainers as a proxy to identifying features important to achieving class change. 

\subsection{Evaluation of Actionable Feature Discovery}
\label{sec:res-afd}

Table~\ref{tbl:sparsity} provides a comparison of five \algo configurations with the DiCE counterfactual algorithm.
Here RND is a baseline counterfactual explanation comparator where actionable features are selected randomly instead of being informed by feature relevance weights.
Results suggests that \algo in the QRel setting achieves the best performance on the Moodle, Alcohol and Income datasets and DiCE achieves best performance on the Loan-2015 dataset (see bold font). \algo using LIME relevance explainer achieves best performance on the Alcohol, Credit and Income datasets while SHAP matches the performance on the Alcohol and Credit Datasets. SHAP achieved best performance on the Moodle dataset.
\algo with QRel and NRel achieve comparable performances across all datasets which resembles findings in Table~\ref{tbl:os}. 
Interestingly, DiCE failed to outperform Random feature ordering on the Alcohol and Income dataset which could be due to the limited amount of features available.

Overall \algo with SHAP and NRel strategy achieves class changes with lowest $\$F$ values (see bold font in Table~\ref{tbl:proximity}) on the Moodle, Income and Credit datasets. 
$\$F$ performance of \algo strategies are better compared to DiCE on all five datasets. 
For instance, for a query in the Loan-2015 dataset, the total amount of change ($\#F \times  \$F$) 
with the DiCE method is $2.01(2.59\times0.7763)$ and with \algo[LIME,NRel] is $0.39(5.51\times0.0711)$. 
In situations where actionable features are not ``easy to change'', it is more feasible to use \algo over DiCE.
With \algo there was no single configuration~(choices for $Rel$ and $p$) that had out performed the rest across datasets. 
It is unusual that \algo with RND resulted in lowest $\$F$ on the Loan-2015 and Alcohol datasets. 
Accordingly, the choice of relevance explainer~(i.e. LIME or SHAP) and ordering strategy~(i.e. QRel or NRel) are seemingly dataset dependent.

\subsection{\algo Counterfactuals}
\begin{figure*}[t]
\centering
\includegraphics[width=0.9\textwidth]{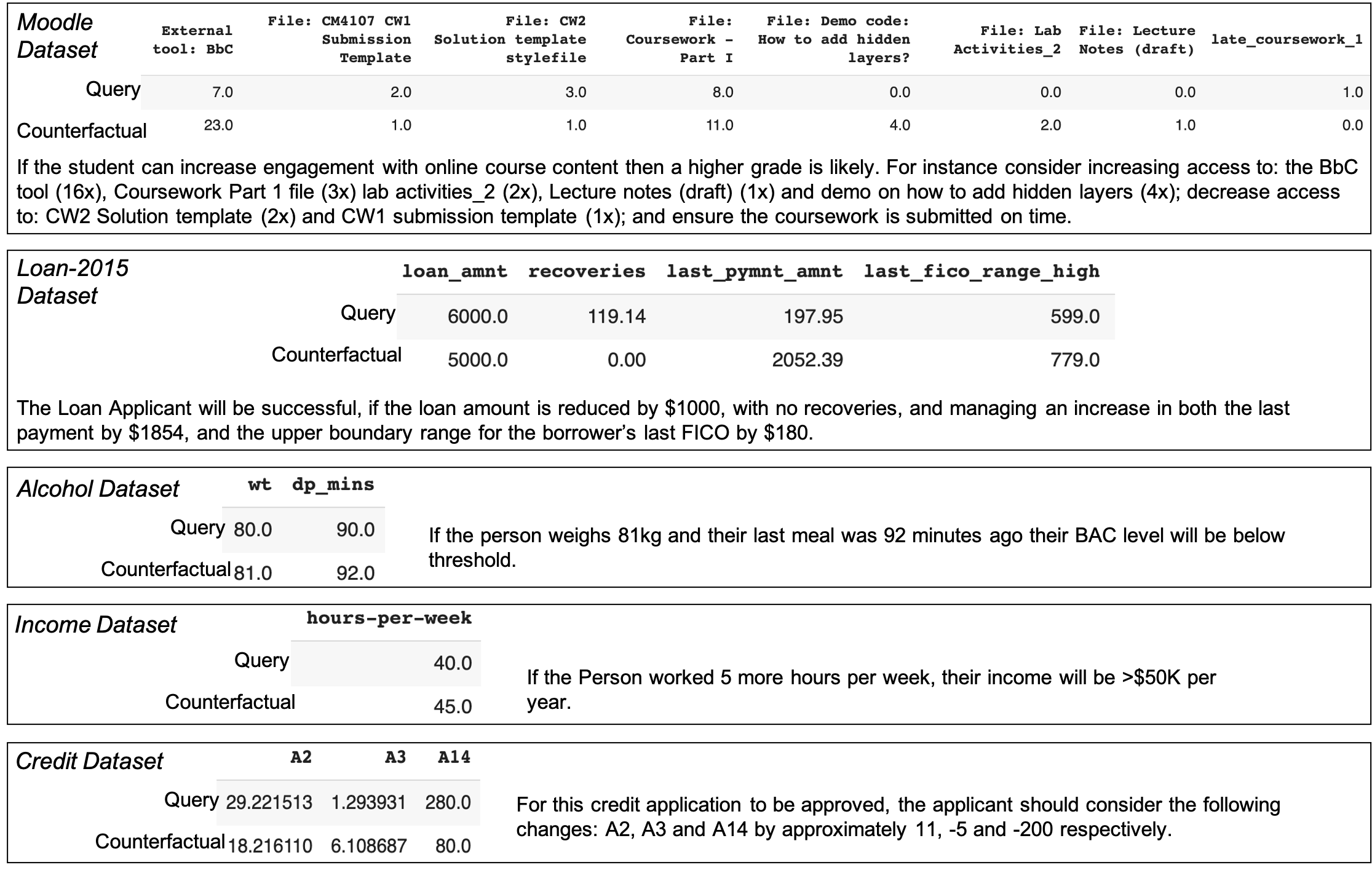}
\caption{Example Counterfactuals} 
\label{fig:examples}
\end{figure*}

Figure~\ref{fig:examples} shows how \algo can be used to form counterfactual textual explanations using an example query from each dataset. 
For purposes of illustration, we selected queries that had a negative outcome and are good candidates to demonstrate actionable feature changes to achieve a desirable class change. 
Here only those actionable features discovered using \algo are shown (and the other features including those with identical values are not). 
A template-based textual explanation generated from the counterfactual is also presented for each example. 
It is interesting to note that with all datasets, \algo is identifying important relationships such as that between a person's weight, the time since the last meal, and the BAC level in the Alcohol Dataset or the relationship between working hours and salary in the income Dataset. 
Although some of these examples are genuine causal relationships, others are relationships which do not directly affect each others values. For instance in the Moodle dataset we observe increased access to learning materials is being related to a positive outcome; however this does not naturally translate to a causal relationship. 
Understanding the types of relationships that are discovered and using those to guide language generation templates remains an important open-area of research for the future. 

\section{Conclusion} 
\label{sec:conc}
In this paper, we presented a novel approach to finding actionable knowledge when constructing an explanation using a counterfactual. 
We used feature relevance explainers as a strategy to discover features that are most significant to a predicted class and then used that knowledge to discover the actionable features to achieve class change with minimal change. 
We demonstrated our approach \algo using five datasets one of which~(Moodle Dataset) is an original contribution. 

Our empirical results showed that SHAP is the most optimal feature relevance explainer for ordering actionable features. 
\algo with QRel and NRel counterfactual methods introduced in this paper have either outperformed or achieved comparable performance over DiCE. 
The results have also highlighted the need to find balance between the number of feature changes and amount of feature change based on the selected actionable features. 
Accordingly, we find that there is conclusive evidence that feature relevance explainers are an important proxy to discovering actionable features and minimising the changes required. 
Future work will expand upon our evaluation to include additional real-world datasets and the use of qualitative evaluation through crowd-sourcing techniques.

\bibliographystyle{ieeetr}
\bibliography{ref}

\end{document}